# Development and Validation of ML-DQA – a Machine Learning Data Quality Assurance Framework for Healthcare

**Mark Sendak**  MARK.SENDAK@DUKE.EDU
*Duke Institute for Health Innovation*
*Durham, NC, USA*

**Gaurav Sirdeshmukh**  GAURAV.SIRDESHMUKH@DUKE.EDU
*Duke University*
*Durham, NC, USA*

**Timothy Ochoa**  TIMOTHY.OCHOA@DUKE.EDU
*Duke University School of Medicine*
*Durham, NC, USA*

**Hayley Premo**  HAYLEY.PREMO@DUKE.EDU
*Duke University School of Medicine*
*Durham, NC, USA*

**Linda Tang**  LINDA.TANG@DUKE.EDU
*Duke University*
*Durham, NC, USA*

**Kira Niederhoffer**  KIRA.NIEDERHOFFER@DUKE.EDU
*Duke University School of Medicine*
*Durham, NC, USA*

**Sarah Reed**  SARAH.REED@JEFFERSON.EDU
*Jefferson Health*
*Philadelphia, PA, USA*

**Kaivalya Deshpande**  KAIVALYA.DESHPANDE@DUKE.EDU
*Duke Health*
*Durham, NC, USA*

**Emily Sterrett**  EMILY.STERRETT@DUKE.EDU
*Duke Health*
*Durham, NC, USA*

**Melissa Bauer**  MELISSA.BAUER@DUKE.EDU
*Duke Health*
*Durham, NC, USA*

**Laurie Snyder**  LAURIE.SNYDER@DUKE.EDU
*Duke Health*
*Durham, NC, USA*

**Afreen Shariff**  AFREEN.SHARIFF@DUKE.EDU
*Duke Health*
*Durham, NC, USA*

**David Whellan**  DAVID.WHELLAN@JEFFERSON.EDU
*Jefferson Health*
*Philadelphia, PA, USA*

**Jeffrey Riggio**  JEFFREY.RIGGIO@JEFFERSON.EDU
*Jefferson Health*
*Philadelphia, PA, USA*

**David Gaieski**  DAVID.GAIESKI@JEFFERSON.EDU
*Jefferson Health*
*Philadelphia, PA, USA*

**Kristin Corey**  KRISTIN.COREY@DUKE.EDU
*Duke Health*
*Durham, NC, USA*

**Megan Richards**  MEGAN.RICHARDS@DUKE.EDU
*Duke University*
*Durham, NC, USA*

**Michael Gao**  MICHAEL.GAO@DUKE.EDU
*Duke Institute for Health Innovation*
*Durham, NC, USA*

**Marshall Nichols**  MARSHALL.NICHOLS@DUKE.EDU
*Duke Institute for Health Innovation*
*Durham, NC, USA*

**Bradley Heintze**  BRADLEY.HEINTZE@DUKE.EDU
*Duke Institute for Health Innovation*
*Durham, NC, USA*

**William Knechtle**  WILLIAM.KNECHTLE@DUKE.EDU
*Duke Institute for Health Innovation*
*Durham, NC, USA*

**William Ratliff**  WILLIAM.RATLIFF@DUKE.EDU
*Duke Institute for Health Innovation*
*Durham, NC, USA*

**Suresh Balu**  SURESH.BALU@DUKE.EDU
*Duke Institute for Health Innovation*
*Durham, NC, USA*

## Abstract

The approaches by which the machine learning and clinical research communities utilize real world data (RWD), including data captured in the electronic health record (EHR), vary dramatically. While clinical researchers cautiously use RWD for clinical investigations, ML for healthcare teams consume public datasets with minimal scrutiny to develop new algorithms. This study bridges this gap by developing and validating ML-DQA, a data quality assurance framework grounded in RWD best practices. The ML-DQA framework is applied to five ML projects across two geographies, different medical conditions, and different cohorts. A total of 2,999 quality checks and 24 quality reports were generated on RWD gathered on 247,536 patients across the five projects. Five generalizable practices emerge: all projects used a similar method to group redundant data element representations; all projects used automated utilities to build diagnosis and medication data elements; all projects used a common library of rules-based transformations; all projects used a unified approach to assign data quality checks to data elements; and all projects used a similar approach to clinical adjudication. An average of 5.8 individuals, including clinicians, data scientists, and trainees, were involved in implementing ML-DQA for each project and an average of 23.4 data elements per project were either transformed or removed in response to ML-DQA. This study demonstrates the importance role of ML-DQA in healthcare projects and provides teams a framework to conduct these essential activities.

## 1. Introduction

The machine learning (ML) in healthcare and clinical research communities have entirely different approaches to developing new products using real-world data (RWD). Until the last decade, nearly all data used for the regulatory approval of novel therapeutics and medical devices was manually curated through high-cost clinical trials. The broad adoption of electronic health records (EHRs) after the HITECH act of 2009 hastened the digitization of patient data and laid the groundwork for new opportunities to leverage EHR data. The 21st Century Cures Act of 2016 formally promoted the use of RWD, including EHR data, medical claims, billing data, registry data, and patient-generated data, in regulatory reviews of products by the Food and Drug Administration (FDA).[1] However, despite increased use of EHR data to assess product safety and efficacy, the FDA continues to closely scrutinize the use of RWD in clinical investigations.[2]

While the clinical research community has approached use of RWD with great trepidation, the ML community zealously develops new algorithms on the same types of datasets. The ML community frequently calls for the open release of new RWD datasets and datasets that are publicly available attract thousands of teams to develop and validate new algorithms.[3] The ML community recognizes that the current approach to commoditize benchmark datasets is untenable.[4] A recent review succinctly described the problem: "It is essential that researchers, advocacy groups, and the public at large understand both the contents of the datasets and how they affect system performance. In particular, as the field has focused on benchmarks as the primary tool for both measuring and driving research progress, understanding what these benchmarks measure (and how well) becomes increasingly urgent."[4] Beginning in 2018, efforts to improve dataset documentation, including Datasheets for Datasets,[5] have emerged and are broadly cited. However, they remain poorly adopted by ML in healthcare teams and they do not account for the rich literature and best practices surrounding RWD use in clinical research.

The current study seeks to close this gap. We build upon seminal work performed by the Observational Health Data Science and Informatics (OHDSI) and Patient Centered Outcomes Research Institute (PCORI) to harmonize a data quality framework for secondary use of EHR data.[6] The PCORI harmonized data quality terminology and definitions were "explicitly scoped to focus on the broad-based evaluation of a large data set typically found in clinical data sharing networks" and were designed to be agnostic of EHR system.[6] The effort convened representatives from 20 of the largest United States distributed research networks along with international representatives and the participants collectively gather data on over 540 million patients. This study also builds upon prior work to adapt the PCORI harmonized data quality framework to assess quality of data curated by a custom-built automated EHR data pipeline.[7] To our knowledge, this study is the first of its kind to develop and validate a data quality assurance (DQA) framework grounded in clinical research RWD best practices on a portfolio of ML projects.

This study has two main objectives. The primary objective is to validate a DQA framework applied to five ML projects across different product development stages, geographic settings, and disease conditions. While conducting this validation, we highlight generalizable practices across projects. The secondary objective is to characterize the resource requirements and effect of utilizing the ML-DQA framework on each ML project.

### 1.1 Generalizable Insights

This study presents a DQA framework validated on five ML in healthcare projects at different stages, in different geographies, and different disease conditions. We distill findings into a set of practices and templates that can easily be adopted by other groups working with RWD to develop ML products in healthcare.

## 2. Methods
### 2.1 Setting

This study is led by an interdisciplinary team at Duke Health. Since the founding of the Duke Institute for Health Innovation (DIHI) in 2013, DIHI has completed over 90 projects, including the development and integration of 15 ML products into clinical care. DIHI's approach to ML product development is heavily influenced by close and longstanding collaborations with leaders of the Duke Clinical Research Institute (DCRI). DCRI works closely with nearly all major therapeutic and device developers to advance novel products through regulatory approval and post-market surveillance. DCRI also serves as the data coordinating center for PCORI and utilizes the harmonized DQA framework to aggregate data from sites across the distributed research network.

A secondary site involved in the study is Jefferson Health, which also conducts a portfolio of clinical research studies with industry sponsors. Duke Health and Jefferson Health are collaborating to expand their clinical research partnership activities to include the development and validation of novel ML products. The current study includes the first time an ML product built at Duke Health undergoes external validation at Jefferson Health. Notably, both Duke Health and Jefferson Health use Epic Systems as their EHR.

### 2.2 Project Selection

The current study includes 5 ML projects. Project details are listed in **Table 1.** Four projects apply the ML-DQA framework to EHR data from Duke Health and one project applies the framework to EHR data from Jefferson Health. The projects focus on predicting surgical complications, pediatric sepsis, adult sepsis, chemotherapy associated adverse events, and maternal morbidity and mortality. The five included projects also represent two different stages of ML product development. The Jefferson Health Sepsis project is an external validation of a previously published algorithm.[8,9] The other four projects are developing new ML algorithms using Duke Health EHR data. These four projects were all selected through a competitive, annual request for applications in which Duke Health senior leaders specify strategic priorities for the organization and front-line clinical staff propose projects aligned with the strategic priorities.

Clinical project leaders were all front-line clinicians who were perceived by Duke Health senior leaders to be well positioned to champion development and adoption of the new ML product. The projects were all undergoing DQA simultaneously between November 1, 2021 and March 31, 2022. All projects have completed the ML-DQA process. All projects have individual IRB approval for data curation and analyses.

**Table 1.** Descriptions of all 5 ML projects

| | **Project Description** | **Outcome of Interest** | **Setting** | **Project Stage** |
|---|---|---|---|---|
| **Pediatric Sepsis** | Develop an ML model to predict sepsis in the pediatric emergency department. The model is built to be used after pediatric patients present to the ED up until time of death, discharge, or sepsis. The model will be used to identify patients who may need treatment for sepsis. | Pediatric Sepsis (see Supplemental Table 1) | Duke Health | Initial ML model development |
| **Lung Transplant** | Develop an ML model to identify patients who are at high risk of poor outcomes after a lung transplant. The model is built to be used after patients are listed for transplant leading up until the time of lung transplant. The model will be used to identify patients who may need pre-operative interventions to minimize risk of post-operative complications. | Mortality at 1 year post-transplant and "Textbook outcomes"[10] | Duke Health | Initial ML model development |
| **Jefferson Sepsis** | Externally validate the Sepsis Watch model built at Duke Health on EHR data curated at Jefferson Health. The model is built to be used after adult patients present to the ED up until time of death, admission, discharge, or sepsis. The model will be used to identify patients who may need treatment for sepsis. | Sepsis as defined by Sepsis Watch | Jefferson Health | External validation of previously developed Sepsis Watch ML algorithm |
| **Immune Related Adverse Events (irAE)** | Develop an ML model to predict which cancer patients treated with immune checkpoint inhibitors are at high risk of immune related adverse events. The model will be used by outpatient oncologists to identify patients who need closer monitoring or additional support to prevent emergency department visits or hospital admissions. | Hospital admissions and Emergency Department visits | Duke Health | Initial ML model development |
| **Maternal Early Warning Score (MEWS)** | Develop an ML model to predict maternal morbidity and mortality. The model is built to be used after pregnant patients present to the hospital up until the time of death, discharge, or adverse event. The model will be used to identify patients who may need treatment. | Hemorrhage, Sepsis, Acute Heart Failure, Acute Respiratory Distress Syndrome, Pulmonary Embolism, Eclampsia, Disseminated Intravascular Coagulopathy, Acute Renal Failure | Duke Health | Initial ML model development |

## 2.3 Machine Learning – Data Quality Assurance (ML-DQA) Framework

The ML-DQA framework focuses on a narrow portion of the data curation process and applies most specifically to ML products developed using structured EHR data. The ML-DQA framework is completed before ML models are developed and does not depend on the model training methodology. Two activities are completed before application of the ML-DQA framework. First, clinical project leaders identify relevant data elements to include in the ML model. Potential data elements are identified from prior literature as well as surveys distributed to clinicians involved in the diagnosis and treatment of the condition of interest. An example project survey is included in the Supplement. Second, data elements selected for inclusion are mapped to the EHR data source and extracted into a project-specific data store. After the queries are refined to ensure capture of all relevant data, we proceed with the ML-DQA framework. The framework has three phases, illustrated in a simplified workflow process diagram in Figure 1.

**Figure 1.** Workflow process diagram describing ML-DQA framework

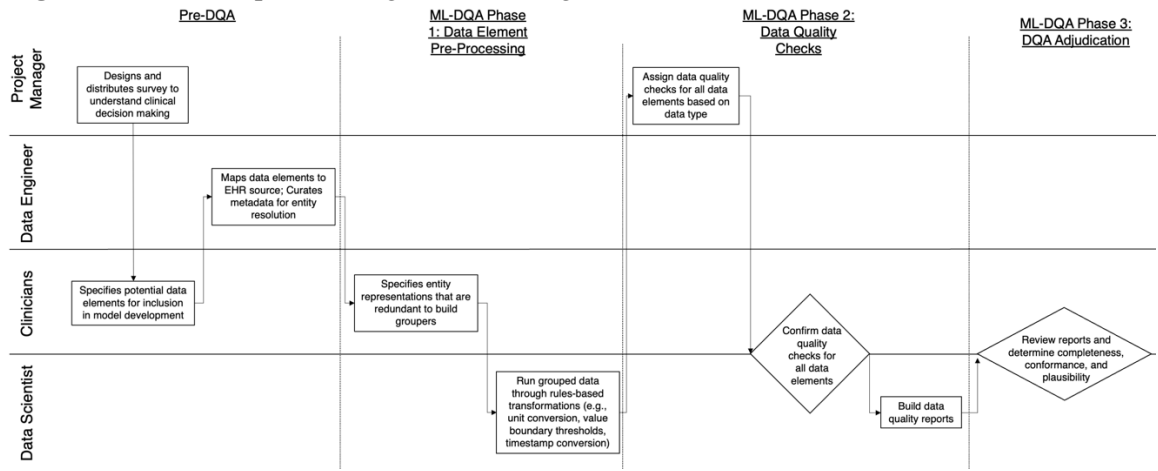

*Phase 1: Data Element Pre-Processing*

    The first step is entity resolution, also called grouping. Entity resolution is required for many different types of healthcare data and can draw upon publicly curated resources, but also often requires local clinical expertise and input. Entity resolution mappings are built for many types of healthcare data, such as laboratory measurements, vital sign measurements, medications, and escalation of care events. For example, glucose is measured dozens of different ways, including blood measurements, urine measurements, measurements taken from point-of-care devices, measurements during fasting, measurements processed at Labcorp, measurements processed at Quest Diagnostics, and measurements taken from both venous and arterial samples. Defining a serum glucose data element grouper requires clinicians to specify a mapping that resolves the redundancy to map various forms of measurement to the same object. A serum glucose grouper is included in the supplement illustrating 38 different ways that glucose was captured in the Duke Health EHR between June 1, 2014 and November 29, 2020. Similar mappings are required for the different ways heart rate is measured, the myriad types of antibiotics that can be administered, and transitions from a general inpatient unit to any of the different types of intensive care units.

    The second step is rules-based data transformations that address common problems encountered with EHR data. These rules sets include unit normalizations to transform various units of measurement into a single reference unit per data element. Threshold-based rules are often developed when no reference unit is available. For example, a substantial proportion of weight measurements are captured in ounces. Weight measurements over 1,200 are assumed to be measured in ounces and are converted to lbs. There are also rules sets designed specifically for the EHR configuration at Duke Health: invalid numeric values are replaced with the value 999999; missing entries for birth date are imputed with the value December 31, 1841; and timestamps are most often recorded in local time and need to be converted to UTC. The rules set library is dynamic and expands over time as new transformations are identified.

*Phase 2: ML-DQAChecks*

ML-DQA for HealthcareThe first step is to assign checks to every data element that assess for completeness, conformance, and plausibility. This step draws upon the PCORI harmonized data quality framework, which defines similar domains of quality.[6] Conformance checks test that variable type, value, range, and computational output match prespecified expectations. Completeness checks report missingness of data and plausibility checks test data variable credibility within clinical contexts. All data elements are assessed in an atemporal and temporal fashion to ensure that data quality is stable over time. Temporal shifts or drifts in data quality are particularly important for ML projects, because model performance can deteriorate if data or the population changes.[11] Checks are defined by clinicians and data scientists, with plausibility checks requiring the most clinical expertise. For example, we try to reproduce associations between known risk factors and an outcome (e.g., older patients have higher rates of in-hospital mortality). Similarly, we try to reproduce heuristics that associate certain interventions with certain clinical factors (e.g., patients who are hypotensive often receive vasopressors).

Once clinicians and data scientists agree on the ML-DQA checks, data quality reports are generated for categories of data elements, including encounters, laboratory measurements, vital sign measurements, and medications. Data quality reports are currently generated using Jupyter notebooks and contain numerous tables and visualizations. The ML-DQA reports are meant to enable rapid assessment of completeness, conformance, and plausibility of each data element.

*Phase III: ML-DQA Adjudication*

The final step in the ML-DQA framework is adjudication of data quality by clinicians. This last step requires project stakeholders to confirm that each data element is fit-for-use in downstream analyses and ML model development. For a data element to be fit-for-use, the reviewer must confirm that the data element is sufficiently complete, conformant, and plausible. These determinations are made by reviewing the ML-DQA Jupyter notebooks and completing a spreadsheet form. An example adjudication form for the lung transplant project is included in the Supplement.

After ML-DQA adjudication, data elements that are not fit-for-use are reviewed by the data scientist and project manager. An alternate source for the data may be identified, the query sourcing data from the EHR may be adapted, or additional transformations may be needed. Modifications to the data element are presented in a new ML-DQA report, which undergoes another round of adjudication by a clinical reviewer. Ultimately, any data elements that continue to not be fit-for-use are not used in ML model development.

**2.4    Validation Analyses**

The primary objective of this study is to validate the ML-DQA framework applied to five diverse ML projects. To do this, we describe how each project completed each of the ML-DQA steps described above and highlight the components of the framework that were most generalizable across projects. We synthesize our findings and present resources that can be used by teams using EHR data to develop new ML in healthcare products.

**2.5    Impact Analyses**

The secondary objective of this study is to characterize the resource requirements and effect of utilizing the ML-DQA framework on each ML project. To do this, we quantify the personnel involved in completed ML-DQA and measure number of data elements that were deemed to not be fit-for-use as well as the number of data elements that required transformation or were excluded from use in ML model development.

**3.  Results**

ML-DQA for HealthcareIn total, the ML-DQA framework was applied to data curated for 247,536 patients across the five ML projects. Cohort sizes ranged from 742 patients who underwent lung transplantation at one hospital between 2014 and 2020 to 212,531 patients who presented to a large, urban hospital in 2019. Cohorts varied by age, including one pediatric cohort (mean age = 7.85 years), a cohort of young women admitted to labor and delivery (mean age = 29.8 years), and a cohort of older adults undergoing chemotherapy (mean age = 65.46 years). Population characteristics for the five ML projects are presented in **Table 2.**

**Table 2.** Cohort characteristics for all ML projects

|  | **Pediatric Sepsis** | **Lung Transplant** | **Jefferson Sepsis** | **Immune Related Adverse Events** | **MEWS** |
|---|---|---|---|---|---|
| Patients, n | 10,492 | 719 | 212,531 | 3,962 | 19,832 |
| Number of encounters, n | 17,491 | 742 | 484,448 | 130,265 | 15,634 |
| Age, y, mean ± SD | 7.85 ± 6.00 years | 55.69 ± 15.19 years | 57.88 ± 18.26 years | 65.46 ± 11.8 years | 29.80 ± 6.08 |
| Female sex in patient population (n%) | 5,140 (49%) | 304 (40.97%) | 119,054 (59.87%) | 1,564 (39.47%) | 19,832 (100%) |
| **Race (n%)** | | | | | |
| ≥2 races | 418 (3.98%) | 8 (1.08%) | 930 (0.44%) | 56 (1.41%) | 1,557 (9.96%) |
| American Indian or Alaska Native | 115 (1.09%) | 3 (0.40%) | 1,292 (0.61%) | 10 (0.25%) | 54 (0.35%) |
| Black or African American | 3232 (30.80%) | 72 (9.70%) | 61,713 (29.07%) | 700 (17.67.%) | 4,826 (30.87%) |
| Caucasian/white | 4971 (47.37%) | 637 (85.85%) | 110,064 (51.85%) | 2,939 (74.18%) | 6,109 (39.08%) |
| Asian | 264 (2.51%) | 4 (0.54%) | 15,377 (7.24%) | 64 (1.62%) | 1,160 (7.42%) |
| Native Hawaiian or Pacific Island | 31 (0.29%) | 1 (0.13%) | NA | 7 (0.18%) | 0 (0%) |
| Other/missing | 1461 (13.92%) | 17 (2.29%) | 22,897 (10.79%) | 194 (4.90%) | 1,928 (12.33%) |
| **Ethnicity (n%)** | | | | | |
| Non-Hispanic | 8590 (81.87%) | 697 (93.94%) | 180,983 (85.26%) | 3647 (98.78%) | 12,127 (77.6%) |
| Hispanic | 1343 (12.80%) | 17 (2.29%) | 13,977 (6.58%) | 55 (1.39%) | 3,221 (20.6%) |
| Other/missing | 544 (5.18%) | 23 (3.10%) | 17,313 (8.16%) | 260 (7.04%) | 286 (1.98%) |
| Cohort dates | **Start Date:** 11/1/2016 **End Date:** 12/31/2020 | **Start Date:** 07/06/2014 **End Date:** 11/29/2020 | **Start Date:** 01/01/2019 **End Date:** 12/31/2019 | **Start Date:** 04/01/2016 **End Date:** 07/31/2021 | **Start Date:** 01/15/2015. **End Date:** 06/01/2020 |
| Exclusion/Inclusion criteria for cohort | **Inclusion:** 30 days < Age < 18 years; Inpatient/ED at DUH, 11/1/2016 - 12/31/2020. **Exclusion:** patients who received care while in L&D departments at any time during their encounter | **Inclusion:** All patients >= 18 y.o. who have had at least one encounter with the lung transplant clinic and a lung transplant procedure. **Exclusion:** <18 y.o | **Inclusion:** >18 yrs, inpatient | **Inclusion:** Age ≥18 years old, 1st immunotherapy administration at Duke Health hospitals between 04/1/2016 – 07/31/2021 **Exclusion:** <18 years old | **Inclusion:** In OB Triage//Post Partum or L&D Triage/Post Partum, non male, older than 1 year old, less than 50 years old, has admission time value. |

All five projects successfully applied the ML-DQA framework described in the methods. Completed steps include: entity resolution and rules-based transformations; ML-DQA check assignment and data quality report generation; and ML-DQA adjudication involving manual review of data element completeness, conformance, and plausibility by clinicians. Results of these activities across all five projects are captured in **Table 3.** Across the five projects, a total 220 new project-specific groupers were built, 2,999 ML-DQA checks were run, and 24 data quality reports were generated.

**Table 3.** Results of data element pre-processing and ML-DQA checks across projects

|  | **Pediatric Sepsis** | **Lung Transplant** | **Jefferson Sepsis** | **Immune Related Adverse Events** | **MEWS** |
|---|---|---|---|---|---|
| *Phase I: Data Element Pre-Processing* | | | | | |
| Pre-existing groupers | 108 | 109 | 30 | 39 | 310 |
| Project-specific groupers | 73 | 35 | 59 | 41 | 12 |
| *Phase II: ML-DQA Checks* | | | | | |
| Number of data quality checks | 389 | 432 | 267 | 1,034 | 877 |
| Number of completeness checks | 144 | 144 | 70 | 508 | 404 |
| Number of conformance checks | 122 | 144 | 132 | 225 | 69 |
| Number of plausability checks | 123 | 144 | 61 | 301 | 404 |
| Number of data quality reports | 4 | 4 | 4 | 7 | 5 |
| Names of data quality reports | Analytes, Encounters, Flowsheets, Medications | Analytes, Encounters, Flowsheets, Medications | Analytes, Encounters, Flowsheets, Medications | Analytes, Flowsheets, Medications, Demographics, Comorbidities, Orders, Encounters | Orders, Analytes, Flowsheets, Medications, Comorbidities |

### 3.1 Generalizable Practices

Five practices emerged that were generalizable across all projects. First, all projects utilized a common set of metadata when conducting data element entity resolution to build groupers. Specifically, there was consensus across projects of the relevant metadata required to group analytes/laboratory measurements, medication administrations, and flowsheet/vital measurements. For analytes/laboratory measurements, the following metadata was curated: component id, component name, component count, common name, test name, procedure name, order name, component numeric value count (frequency of component), component numeric mean, top specimen source values, top specimen source counts, top reference unit values, top reference unit counts, normal upper bound, and normal lower bound. For medication administrations, the following metadata was curated: medication administration record (MAR) medication count, raw MAR name, top MAR action label values, top MAR action label counts, top route label values, top route label counts, top raw volume dose values, top raw volume dose counts, top raw volume dose unit values, and top raw volume dose unit counts. Lastly, for flowsheets/vital measurements, the following metadata was curated: flowsheet measure id, flowsheet measure name, flowsheet measure name count, display name, value type label, top flowsheet measure values, and top flowsheet measure value counts.

The second generalizable practice was the development of automated utilities to build default groupings for medical comorbidities and medication therapeutic classes. Rich public ontologies exist for both these domains of healthcare data. Historical ICD codes are mapped to AHRQ clinical classification software comorbidities, incorporating findings from prior research comparing different approaches to grouping ICD codes.[12] Similarly, all raw medication names are parsed, tokenized, and then sent to the RxNorm API to retrieve potential tags mapping each medication to therapeutic classes.

The third generalizable practice was the collection of rules-based transformations into a local library and set of software scripts. After applying the ML-DQA framework to the five projects, 22 categories of rules-based transformations emerged. **Table 4** contains a complete list of the rules-based transformations and maps each different type of data element to the relevant set of transformations. Example transformations include normalizing units that are labeled, applying thresholds to result values to impute missing units, parsing result values to remove indeterminate results, and detecting the bounds at which instruments reliably report measurement values. Project-specific and general transformations are now stored in a centralized library.

**Table 4.** Unified approach for mapping data elements to rules-based transformations

| | **Rules-Based Transformations** | **Example** |
|---|---|---|

*Analyte/Laboratory Measurement*

| | **Rules-Based Transformations** | **Example** |
|---|---|---|
| Numeric | •Unit normalization performed through reference unit mapping<br>•Parse specimen source value and subset by source<br>•Parse typos and non-ASCII characters included in unit of measure<br>•Apply project-specific or general upper and lower bound<br>•Parse string text that conveys numeric result is outside range of measurement instrument and replace with upper bound<br>•Ensure time stamp is captured in UTC | •Normalize serum creatinine values reported in mg/dL and mg/mL<br>•Identifying and dropping serum glucose values that have specimen source of urine<br>•Find and replace greek letters and typos used in units and replace with ASCII characters<br>•Remove serum creatinine values over 150 mg/dL<br>•Convert point-of-care glucose value of ">600 mg/dL" to "600 mg/dL"<br>•Convert timestamp from EDT to UTC |
| Categorical | •Parsing specimen source value and subsetting by source<br>•Map character string to hierarchical value through expert-derived reference table<br>•Map character string to binary value through expert-derived reference table<br>•Parse result text to identify indeterminate values<br>•Ensure time stamp is captured in UTC | •Identify and drop blood cultures that have specimen source pleural fluid<br>•Map blood culture results to negative, likely contaminant, and likely pathogen<br>•Map HIV antibody titer levels to positive or negative result<br>•Replace "hemolyzed sample" with missing value<br>•Convert timestamp from EDT to UTC |

*Flowsheet/Vital Sign Measurement*

| | **Rules-Based Transformations** | **Example** |
|---|---|---|
| Numeric | •Apply project-specific or general upper and lower bound<br>•Parse string value to generate numeric<br>•Unit normalization performed through reference unit mapping<br>•Unit normalization through expert-derived thresholds<br>•Ensure time stamp is captured in UTC | •Remove pulse values over 400 beats per minute<br>•Transform blood pressure (120/80) to separate systolic (120) and diastolic (80)<br>•Convert kg weight values to lbs<br>•Temperature values above 60 are assumed to be Fahrenheit<br>•Convert timestamp from EDT to UTC |
| Categorical | •Map character string to hierarchical value through expert-derived reference table<br>•Map character string to binary value through expert-derived reference table<br>•Map character string with multi-select to hierarchical value through expert-derived reference table<br>•Ensure time stamp is captured in UTC | •Ranking oxygen support such that room air < nasal canula < CPAP < mechanical ventilation<br>•Documentation of impella device in flowsheet represented as 1<br>•If patient is documented to be both alert and oriented and sedated, take minimum level of consciousness<br>•Convert timestamp from EDT to UTC |

*Medication Administrations*

| | **Rules-Based Transformations** | **Example** |
|---|---|---|
| Categorical | •Parse MAR action name and subset by action<br>•Ensure time stamp is captured in UTC | •Identify and drop antibiotic administrations where the MAR action name is "held"<br>•Convert timestamp from EDT to UTC |

The fourth generalizable practice was a unified approach to assign ML-DQA checks to each type of data element. **Table 5** illustrates the approach to map different data element types to corresponding conformance, completeness, and plausibility checks. For example, each numeric analyte or laboratory measurement is assigned the following completeness checks: report total number of measurements across cohort, report proportion of patients with at least 1 measurement, and generate a histogram of number of measurements per patient. By standardizing ML-DQA check assignment, this also allowed teams across projects to reuse code to generate the data quality reports. For example, the first ML project to generate data quality reports was the pediatric sepsis project. All subsequent projects adapted the code after assigning checks using the framework in **Table 5**. Data elements that were not analytes/laboratory measurements, vital sign measurements/flowsheets, medications, or encounter-level patient characteristics required more customized quality checks. An example section from a flowsheet report for the lung transplant project is included in the Supplement.

**Table 5.** Unified approach for mapping data elements to ML-DQA checks

| | **Associated Meta-data** | **Conformance Checks** | **Completeness Checks** | **Plausability Checks** |
|---|---|---|---|---|
| *Analyte/Laboratory Measurement* | | | | |
| Numeric | Order time, collection time, result time, specimen source, reference unit | •Reference unit frequency table<br>•Specimen source frequency table<br>•Table showing deciles of values | •Total number of measurements across cohort<br>•Proportion of patients with at least 1 measurement<br>•Histogram of number of measurements per patient | •Number of measurements per month across cohort<br>•Box plot of measurement value distribution per month across cohort |
| Categorical | Order time, collection time, result time, specimen source, reference unit | •Reference unit frequency table<br>•Specimen source frequency table<br>•Frequency table of all string values | •Total number of measurements across cohort<br>•Proportion of patients with at least 1 measurement<br>•Histogram of number of measurements per patient | •Number of measurements per month across cohort<br>•Line plot of number of values per categorical value per month across cohort |
| *Flowsheet/Vital Sign Measurement* | | | | |
| Numeric | Measurement time | •Table showing deciles of values | •Total number of measurements across cohort<br>•Proportion of patients with at least 1 measurement<br>•Histogram of number of measurements per patient | •Number of measurements per month across cohort<br>•Box plot of measurement value distribution per month across cohort |
| Categorical | Measurement time | •Frequency table of all string values | •Total number of measurements across cohort<br>•Proportion of patients with at least 1 measurement<br>•Histogram of number of measurements per patient | •Number of measurements per month across cohort<br>•Line plot of number of values per categorical value per month across cohort |
| *Medication* | | | | |
| Categorical | Administration time, order time, MAR action label, route | •MAR action name frequency table<br>•Frequency table of all raw medication names | •Total number of med administrations across cohort<br>•Proportion of patients with at least 1 measurement<br>•Histogram of number of administrations per patient | •Number of med administrations per month across cohort<br>•Box plot of number of med administrations per patient per month across cohort |
| *Encounter-Level* | | | | |
| Numeric | ED arrival time, admission time, discharge time | •Table showing deciles of values | •Total number of values across cohort<br>•Proportion of patients with a value<br>•Histogram of number of values per encounter | •Number of encounter values per month across cohort<br>•Box plot of encounter value distribution per month across cohort |
| Categorical | ED arrival time, admission time, discharge time | •Frequency table of all string values | •Total number of values across cohort<br>•Proportion of patients with a value<br>•Histogram of number of values per encounter | •Number of measurements per month across cohort<br>•Line plot of number of values per categorical value per month across cohort |

The fifth generalizable practice was the development of a unified approach to adjudicate the ML-DQA reports. Each adjudication form had 4 sections and was expected to be completed by both a data scientist and a clinician working on the project, as depicted in **Figure 1**. Clinical trainees working full-time as data scientists through the DIHI Clinical Research and Innovation Scholarship were target users.[13] The first section of the adjudication form included relevant metadata and aggregate data from a related dataset. For example, an adjudication form for numeric analytes included frequency of measurement, mean, standard deviation, and reference unit drawn from a previously curated dataset of individuals in the same geography and the same age. These reference values can be obtained either through prior work or through published literature using a similar cohort. The second section of the adjudication form asked the reviewer to respond yes or no to three questions: is the data sufficiently complete? Is the data sufficiently conformant? Is the data sufficiently plausible? The reviewer could also enter notes for each of the three domains of data quality. The third section of the adjudication form asked the reviewer to answer yes or no to six questions: Do we include the data element? Do we transform the data element? Are there data element values we exclude? Which values do we exclude from the data element? Is there a lower bound for inclusion? Is there an upper bound for inclusion? The fourth and final section of the adjudication form allowed the reviewer to write questions or comments in free text. An example adjudication form is included in the Supplement.

### 3.2   ML-DQA Personnel and Capabilities Requirements

The personnel involved in implementing the ML-DQA framework for each ML project are listed in **Table 6.** The mean number of people involved in the process for each project was 5.8 and the median was 5. Each project had at least one clinical expert, one data scientist, one clinical trainee, one clinician data scientist, and one project manager involved in process. The steps that required the most clinician involvement were entity resolution or grouping during phase one, defining the data quality checks during the second phase, and ML-DQA adjudication during the third phase, as illustrated in **Figure 1**. The steps that required the least clinician involvement were the mapping to data elements to the EHR, curation of meta-data for entity resolution, data element pre-processing, and generating the ML-DQA reports.

**Table 6.** Personnel involved in ML-DQA for each ML project

| Project | Personnel Involved | Roles and Capabilities |
|---|---|---|
| Pediatric Sepsis | 5 | data science; clinical expert; project manager; clinical data scientist; clinical trainee |
| Lung Transplant | 8 | clinical expert (x3); project manager (x2); clinical data scientist; clinical trainee (x2) |
| Jefferson Sepsis | 5 | data science; clinical experts (x2); clinical data scientist; clinical trainee |
| Immune Related Adverse Events | 6 | data science (x2); clinical expert; project manager; clinical data scientist; clinical trainee |
| MEWS | 5 | data science; clinical expert; project manager; clinical data scientist; clinical trainee |

### 3.3   ML-DQA Effect

The overall effect of the ML-DQA framework on ML project data is illustrated in **Table 7.** All projects that implemented the ML-DQA framework identified data elements that needed to be transformed or removed from ML model development. The ML-DQA framework surfaced an average of 12.2 data elements per project that needed to be transformed in order to be included in ML model development. Similarly, the ML-DQA framework surfaced an average of 11.2 data elements per project that were deemed not fit-for-use and removed from ML model development.

**Table 7.** Effect of ML-DQA framework on data used for ML projects

| | Pediatric Sepsis | Lung Transplant | Jefferson Sepsis | Immune Related Adverse Events | Maternal Early Warning Score |
|---|---|---|---|---|---|
| **Data Elements Transformed** | 11 | 9 | 13 | 18 | 10 |
| **Example of Transformed Data Element** | Converting non-numeric value to a mapped numeric value; Capillary_refill has multiple text options which were converted to a binary 1 (abnormal) or 0 (normal) | Removed far outliers including: INR of 0, Na of 5.4, TSH of 3320 and 0.01. Updated grouper for creatinine which mistakenly included a urine lab. Updated grouper for TPO antibodies. | Replaced 744,761 values of 9999999 with NaN; Converted ammonia values from ug/dL to umol/L; Custom entity resolution mappings for oxygen device and level of consciousness | Excluded implausible values (e.g., HR < 20, RR < 5); removed elements of premade groupers when necessary (e.g., urine labs appearing in serum groupers); combined rare elements into broader groupers (ie rare autoantibodies like RNP, smith, SSA/B combined to "autoantibody +") | Replaced values with 999999; Converted units; Broaded inclusion of medication administrations |
| **Data Elements Excluded** | 9 | 22 | 9 | 12 | 4 |
| **Example of Excluded Data Element** | crp_high_sensitivity had very low count of data points and measurement of the data element was not stable over time | Albumin to creatinine ratio was excluded, because only 11% of the cohort had data for this data element. | Data elements having little clinical utility (e.g., 'pH' is more often used than 'arterial pH'); Values that are inherently wrong (e.g., troponin had far fewer counts than expected) | A clinician had identified growth hormone levels as potentially important to our outcomes, however <0.05% of our cohort had this lab. It was removed from analysis. | Certain types of orders, such as head MRI and nasal cannula, because they occurred so infrequently |
| **Most Important Adaptation or Finding** | "Personally, when I was initially working with the peds sepsis data, I thought since the vitals have been through the pre-processing step, there shouldn't be any additional outliers. However, the QA process (especially through visualizations) revealed the need for additional processing (with attention to detail) to make sure the data is ready for modeling. I also realized that it's crucial to work with people with clinical expertise in the QA process." | "Since most projects are working with the same underlying clinical database. We all work with similar variables. Therefore errors in groupers in one project can be shared with other projects. For example, the irAE project uncovered that the creatinine grouper contained urine creatinine values along with serum creatinine values which was incorrectly skewing the mean. Her find allowed me to isolate this error and remove the urine creatinine elements from the creatinine grouper in the Lung Transplant project." | "Not necessarily a specific finding, but leadership and clinicians were not aware of the typical state of a raw dataset nor the process required to clean it. Even clinicians that work in informatics often use analytic or modelling products that do not have transparency regarding the state of the inputs. They take it on faith that the product or vendor has "taken care of" data quality with little oversight. It is often the case that even though the data quality issues are invisible to the user, it doesn't mean it's not there." | "QA helped clarify our thinking on a number of elements. While reviewing graphs of element stability over time we identified discrepancies in the time period over which 2 raw data files were collected. If we hadn't done QA we would not have identified this issue which affected a significant number of downstream elements. QA made us slow down and be methodical about each data type." | "I found that QA highlighted what kinds of data were well represented (comorbidities), versus data like labs are much rarer. I think this is a particular difficulty of time-based modeling - the most definitive/useful data is much less frequently received by our models." |

Data scientists across projects also highlighted specific adaptations or findings that were most salient and informative across projects. A common sentiment was surprise at the state of the raw data and the effort required to make the data usable for ML model development. Several specific findings also cut across projects. Many projects identified lab measurements reported in incorrect units, a grouper for a blood serum analyte included instances measured from urine, and high rates of missingness for certain variables required developing new SQL extracts to source data from alternate tables in the EHR database.

## 4. Discussion

The ML-DQA framework presented in this study made a significant, measurable impact on the quality of data used for every ML in healthcare project. On average, over 23 data elements per project were either transformed or removed altogether for ML model development because of conformance, completeness, or plausibility problems. Unfortunately, these findings are consistent with other results in high-stakes domains.[14] Data cascades, defined as "compounding events causing negative, downstream effects from data issues, resulting in technical debt over time" were found to be widely prevalent: 92% of AI practitioners reported experiencing one or more and 45.3% reported two or more cascades in a given project.[14] The current study further illustrates the massive scale of even isolated conformance problems. For example, the Jefferson Sepsis project contained 744,761 lab measurements with an invalid numeric value 999999.

The current study is situated between three bodies of work that have emerged in the ML community. First, there is literature advocating for much needed culture change surrounding data curation. In 2020, Timnit Gebru and Eun Seo Jo encouraged ML practitioners to embrace lessons from historical archives, including: honoring and rewarding the labor of full-time curators responsible for weighing the risks and benefits of gathering different types of data and theoretical frameworks for appraising collected data; codes of conduct/ethics and a professional framework for data curators; and standardized forms of data documentation.[15] A more recent qualitative study also found that efforts to improve data quality are not rewarded: "Models were reported to be the means for prestige and upward mobility in the field with ML publications that generated citations, making practitioners competitive for AI/ML jobs and residencies. 'Everyone wants to do the model work, not the data work.' Many practitioners described data work as time-consuming, invisible to track, and often done under pressures to move fast due to margins— investment, constraints, and deadlines often came in the way of focusing on improving data quality." [14] This advocacy led NeurIPS to launch a new *Datasets and Benchmarks* track to serve as a venue for work related to the creation of high-quality datasets and discussion of ways to improve dataset development.

The second emerging literature in the ML community relates to dataset documentation. Numerous frameworks have emerged including Datasheets for Datasets,[5] the Dataset Nutrition Label,[16] FactSheets,[17] Data Cards,[18] and Healthsheet.[19] These frameworks include varying types and amounts of information about datasets, but they collectively have numerous shortcomings when it comes to data quality for ML in healthcare. The documentation frameworks assume that the dataset is fixed, and that the data consumer is unable to directly access the source system to address flaws in data collection or data representation. When applying the ML-DQA framework in the current study to the 5 healthcare ML projects, many iterations of queries and transformations were required to optimize data quality. There were even cases where low rates of complete data prompted clinical leaders to advocate for changes in workflow to improve data capture. The previously described data documentation frameworks also assume that the dataset curator has the skills and expertise to assess data quality. For ML in healthcare projects, clinical domain expertise is required to assess data conformance, completeness, and plausibility. For the 5 ML projects included in this study, an average of 5.8 people were involved in conducting DQA, often including multiple clinicians and a data scientist.

The third relevant literature in the ML community relates to tools that support the testing and validation of ML datasets. For example, the Data Linter was built at Google and published as a Python package that identifies miscoding errors in data, outlier values or errors in scaling, and packaging errors that cause duplication or missing values.[20] Many similar tools and frameworks exist and are described in a detailed 2019 review by Zhang et al.[21] However, while many approaches are technically sophisticated, they do not address the core data quality challenges described in this study. The flaws in completeness, conformance, and plausibility addressed by the ML-DQA framework are not adversarial. These are inherent flaws within data generating processes and are not introduced by an outside actor seeking to perturb an ML model. Similarly, these data quality flaws do not represent skew between training and testing. The flaws are pervasive and occur across cohorts, geographies, and disease conditions.

The current study ties together these relevant bodies of work and bridges clinical research best practices to ML in healthcare. Pharmaceutical companies, for example, have long placed a premium on high-quality RWD, demonstrated by the $2 billion acquisition of Flatiron Health by Roche in 2018 and recent investments in ConcertAI ($150M series C), Aetion ($110M series C), and Verana Health ($150M series E).[22] The ML-DQA framework validated in this study brings the same level of rigor to ML in healthcare. We hope this study prompts further dialogue and research to streamline the ML-DQA process and cultivates expertise in the ML for healthcare community. Similarly, we expect the different documentation artifacts generated throughout the ML-DQA framework to help close the accountability gap in ML for healthcare.[23] Rather than advocating for a single dataset label, we present different sets of documentation completed by different individuals with different expertise in an iterative fashion to optimize data quality. The data dictionary, data element groupings, data quality checks, ML-DQA reports, and adjudication files collectively record and log the process. Lastly, we present tables and example reports that can be rapidly implemented for new projects. A team developing a new ML for healthcare model can apply the ML-DQA frameworks in **Table 4** and **Table 5** to map a set of data elements to rules-based transformations and data quality checks to report out to a reviewing clinician. In future work, we hope to further automate the process for performing data element pre-processing, assigning quality checks, and generating quality reports.

### 4.1 Limitations

The current study has several limitations. First, while we embrace a consensus definition of data quality developed by research networks across the United States, data quality is in the eye of the beholder. The domains of completeness, conformance, and plausibility may not capture all aspects of data quality and new projects may need to extend the ML-DQA framework. Similarly, all projects included a single clinical expert and data quality determinations may differ if evaluated through a consensus process engaging a greater number of domain experts. Second, the ML-DQA framework may not generalize to all ML for healthcare model development efforts. We tried to minimize this risk by applying the framework to projects across two sites using cohorts of different ages for different medical conditions, but there may be project-specific challenges and nuances that emerge in new settings. All projects presented in the current study did use structured EHR data sourced from instances of Epic Systems. ML products built using EHR data sourced from a different software vendor or using unstructured notes or images may also require adaptation of the ML-DQA framework. We tried to minimize this risk by building upon the PCORI data quality framework, which was developed to be agnostic to EHR-vendor and has been used on datasets that span all major EHR vendor systems.

Third, while we quantify the number of checks, number of reports, number of people involved, and number of flaws identified and addressed, these metrics do not measure the full costs of implementing ML-DQA or the value created or captured by conducting ML-DQA. We did not account for personnel effort to conduct ML-DQA across projects, although prior work does highlight the significant costs associated with curating data for EHR models.[24] There is growing consensus that data quality labor is important and should be prioritized, but high-quality data may not create immediate value. Pharmaceutical companies expect to reap significant financial rewards from newly approved therapeutics, whereas the return on investment for health systems that validate ML products on high-quality data is much smaller. Incentives to drive adoption of ML-DQA in healthcare are not yet mature.

Fourth, we focused the current study on presenting a comprehensive, generalizable ML-DQA framework, rather than identifying which ML-DQA components are highest yield. Like code coverage with software testing, ML in healthcare teams can check many aspects of data quality, but should not expect to achieve 100% coverage. Unfortunately, our results showed that all ML projects surfaced data quality problems and future analysis will be required to determine if problems are most associated with specific types of data elements. Teams facing significant resource or time constraints may be able to minimize effort by limiting the number of data sources and number of data elements considered for ML model development.

## 5. Conclusion

In conclusion, this study develops and validates the ML-DQA framework for machine learning in healthcare. The ML-DQA framework is applied to five ML projects across two geographies, different medical conditions, and different cohorts. Five generalizable practices emerge: all projects used a similar method to group redundant data element representations; all projects used automated utilities to build diagnosis and medication data elements; all projects used a common library of rules-based transformations, illustrated in **Table 4**; all projects used a unified approach to assign data quality checks to data elements, illustrated in **Table 5**; and all projects used a similar approach to clinical adjudication. An average of 5.8 individuals, including clinicians, data scientists, and trainees, were involved in conducting ML-DQA for each project and across projects an average of 23.4 data elements were either transformed or removed in response to ML-DQA. This study demonstrates the important role of ML-DQA in healthcare projects and provides teams a framework to conduct these essential activities.

**Supplement**

**Table 1.** Pediatric Sepsis Definition

| **Infection** | |
|---|---|
| | Blood culture (ordered or collected) OR transfer from external healthcare facility, AND |
| | Antibiotic administration within 2 days of blood culture or transfer day |
| **Acute organ dysfunction (any criteria below within 1 day of blood culture or transfer)** | |
| Cardiovascular | > 60 mL/kg isotonic fluid boluses within 7 hours, OR |
| | New, additional, or increased dose of vasoactive medication, OR |
| | Blood lactate >= 2.0 mmol/L |
| Respiratory | New invasive/noninvasive mechanical ventilation |
| Hematologic | Platelet < 100,000 cells/uL and >= 50% decline from baseline |
| Kidney | Serum creatinine >= 2x baseline and exceeding threshold for age |

**Clinician Survey for Lung Transplant Project**
1. What variables do you use nearly every time you make the decision to relocate or list a patient for lung transplant? In other words, what recipient information can we not afford to miss during the listing conference? Please list 10 pieces of information (or as many as you can think of) to rule in or rule out listing.
2. What top 5 variables would you want to see presented in a stratification tool for pre lung transplant patients at the time of a listing conference?
3. What donor information is critical pre lung transplant? What donor information must you understand so that you can plan to reduce patient morbidity and mortality? Please list 10pieces of information (or as many as you can think of) to rule in or rule out surgery.
4. After listing and matching a donor, what other preoperative variables do you use to reduce patient morbidity and mortality? Please list 10 pieces of information (or as many as you can think of).
5. What information must we monitor immediately after the lung transplant? Please list 10 pieces of information (or as many as you can think of) to rule in or rule out discharge.
6. In addition to survival, what other outcomes are critical to monitor? Please list 5 pieces of information (or as many as you can think of).
7. What data should we be wary of? What are 5 data elements that, if a model identified as important predictors, would make you skeptical of the model? What data would a novice or outsider think are predictive, but you wouldn't trust as predictive?
8. What aren't we asking? Is there anything else that is important for us to consider as we work to improve the triage and treatment of lung transplant patients?
9. (Optional) – External tools/scores/decision aids review
Please share what you consider to be the 5 most relevant tools that either: (1) help with clinical decision-making to avoid negative outcomes due to lung transplant, (2) describe risk factors for negative outcomes due to lung transplant, or (3) describe protective factors for negative outcomes due to lung transplant